\documentclass[runningheads]{llncs}
\usepackage[T1]{fontenc}
\usepackage{graphicx}
\usepackage{cite}
\usepackage{amsmath,amssymb,amsfonts}
\usepackage{algorithmic}
\usepackage{comment}
\usepackage{textcomp}
\usepackage{subfigure}
\usepackage{hyperref}
\usepackage{xcolor}
\usepackage{float}
\usepackage{multirow}

\begin{document}
\title{Semi-supervised Pathological Image Segmentation via Cross Distillation
of Multiple Attentions}
%

\titlerunning{Semi-Supervised Segmentation via Cross Distillation of Multiple Attentions}
\author{Lanfeng Zhong\inst{1}  \and Xin Liao\inst{2} \and Shaoting Zhang\inst{1,3} \and Guotai Wang\inst{1,3}}

\authorrunning{L. Zhong et al.}

\institute{University of Electronic Science and Technology of China \and Department of Pathology, West China Second University Hospital, Sichuan University \email{xinty927@163.com}\\ }

\institute{University of Electronic Science and Technology of China, Chengdu, China \and Department of Pathology, West China Second University Hospital,  Sichuan University, Chengdu, China \and Shanghai Artificial Intelligence Laboratory, Shanghai, China. \email{guotai.wang@uestc.edu.cn}}
\maketitle              
\begin{abstract}
Segmentation of pathological images is a crucial step for accurate cancer diagnosis.
However, acquiring dense annotations of such images for training is
labor-intensive and time-consuming. To address this issue, Semi-Supervised Learning
(SSL) has the potential for reducing the annotation cost, but it is 
challenged by a large number of unlabeled training images.
In this paper, we propose a novel SSL method based on Cross Distillation 
of Multiple Attentions (CDMA) to effectively leverage unlabeled images.
Firstly, we propose a Multi-attention Tri-branch Network (MTNet) that 
consists of an encoder and a three-branch decoder, with each branch using a different attention mechanism that calibrates features 
in different aspects to generate diverse outputs.
Secondly, we introduce Cross Decoder Knowledge Distillation  (CDKD) between 
the three decoder branches, allowing them to learn from each other's soft labels to 
mitigate the negative impact of incorrect pseudo labels in training.
Additionally, uncertainty minimization is applied to the average prediction of the three branches, which further regularizes predictions on unlabeled images and encourages inter-branch consistency.
Our proposed CDMA was compared with eight state-of-the-art SSL methods on the public DigestPath dataset, and the experimental results showed that our method outperforms
the other approaches under different annotation ratios.
The code is available at \href{https://github.com/HiLab-git/CDMA}{https://github.com/HiLab-git/CDMA.}

\end{abstract}
\section{Introduction}
Automatic segmentation of tumor lesions from pathological images
plays an important role in accurate diagnosis and quantitative evaluation of cancers.
Recently, deep learning has achieved 
remarkable performance in pathological image 
segmentation when trained with a large and 
well-annotated
dataset~\cite{hou2019dual,xie2019deep,shen2020deep}.
However, obtaining dense annotations for pathological images is challenging and 
time-consuming, 
due to the extremely large image size (e.g., 10000 $\times$ 10000 pixels), scattered spatial distribution, and complex shape of lesions.

Semi-Supervised Learning (SSL) is a potential technique to reduce the annotation cost via learning from a limited number of labeled data along with a large amount of unlabeled data.
Existing SSL methods can be roughly divided into two categories: 
consistency-based~\cite{luo2022urpc, tarvainen2017mean,yu2019uncertainty}  and pseudo label-based~\cite{chen2021semi} methods.
The consistency-based methods impose consistency constraints  on the predictions of an unlabeled image under some perturbations. For example, Mean Teacher (MT)-based methods~\cite{tarvainen2017mean,yu2019uncertainty} encourage consistent predictions between a teacher and a student model with noises added to the input. 
Xie et al.~\cite{xie2020pairwise} introduced a pairwise
relation network to exploit semantic consistency between each pair of images in the feature space.
Luo et al.~\cite{luo2022urpc} proposed an uncertainty rectified pyramid consistency between multi-scale predictions.
Jin et al.~\cite{jin2022semi} proposed to encourage the predictions of auxiliary decoders and a main decoder to be consistent under perturbed hierarchical features.
Pseudo label-based methods typically generate pseudo labels for labeled images 
to supervise the network~\cite{Fan2020}. Since using a model's prediction to 
supervise itself may over-fit its bias, Chen et al.~\cite{chen2021semi} 
proposed Cross Pseudo Supervision (CPS) where two networks learn from each 
other's pseudo labels generated by \textit{argmax} of the output prediction. MC-
Net+~\cite{wu2022mutual} utilized multiple 
decoders with different upsampling strategies to obtain slightly different outputs, and each decoder's probability output was sharpened to serve as pseudo labels to supervise the others.
However, the pseudo labels are not accurate and contain a lot of noise, using \textit{argmax} or sharpening operation will lead to over-confidence of potentially wrong predictions,  which limits the performance of the models.
Additionally, some related works advocated the entropy-minimization methods.
Typical entropy Minimization (EM)~\cite{vu2019advent} that aims to reduce the uncertainty or entropy in a system. Wu et al.~\cite{wu2022cross} directly
applied entropy minimization to the segmentation results.

\vspace{-1mm}
In this work, we propose 
a novel and efficient method based on Cross Distillation with
Multiple Attentions (CDMA) for semi-supervised 
pathological image segmentation.
Firstly, a Multi-attention Tri-branch Network (MTNet) is proposed to efficiently obtain diverse outputs for a given input. Unlike MC-Net+~\cite{wu2022mutual} that is based on different upsampling strategies, our MTNet uses different attention mechanisms in three decoder branches that calibrate features in different aspects to obtain diverse and complementary outputs. Secondly, inspired by the observation that smoothed labels are more effective for noise-robust learning found in recent studies~\cite{muller2019does,xu2020fnkd}, we propose a Cross Decoder Knowledge Distillation (CDKD) strategy to better leverage the diverse predictions of unlabeled images. In CDKD,  each branch serves as a teacher of the other two branches using soft label supervision, which reduces the effect of noise for more robust learning from inaccurate pseudo labels than \textit{argmax}~\cite{chen2021semi} and sharpening-based~\cite{wu2022mutual} pseudo supervision in existing methods. Differently from typical Knowledge Distillation (KD) methods~\cite{hinton2015distilling,zhao2022decoupled} that require a pre-trained teacher to generate soft predictions, our method efficiently obtains the teacher and student's soft predictions simultaneously in a single forward pass. In addition, inspired by EM~\cite{vu2019advent}, we apply an uncertainty minimization-based regularization to the average probability prediction across the decoders, which not only increases the network's confidence, but also improves the inter-decoder consistency for leveraging labeled images.

The contribution of this work is three-fold: 1) A novel framework named CDMA based on MTNet is 
introduced for semi-supervised pathological image segmentation, which leverages different 
attention mechanisms for generating diverse and complementary 
predictions for unlabeled images;
2) A Cross Decoder Knowledge Distillation method
is proposed for robust and efficient learning 
from noisy pseudo labels, which is combined with 
an average prediction-based uncertainty 
minimization to improve the model's performance; 
3) Experimental results show that the proposed CDMA
outperforms eight state-of-the-art SSL methods on the public
DigestPath dataset~\cite{da2022digestpath}.

\begin{figure}[t] \centering
\includegraphics[width=\textwidth]{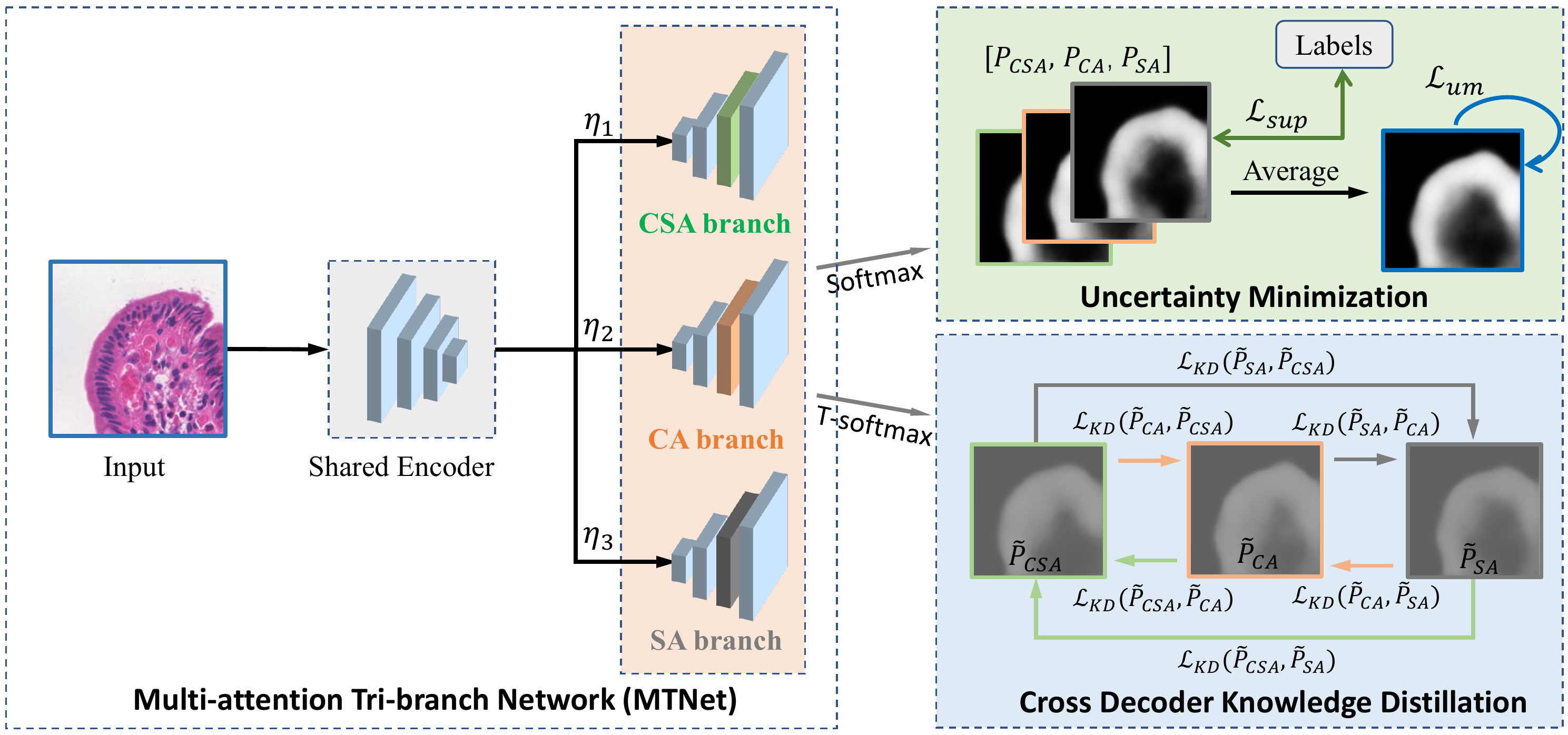}
\caption{Our CDMA for semi-supervised segmentation. Three decoder branches use different attentions to obtain diverse outputs. Cross Decoder Knowledge Distillation (CDKD) is proposed to better deal with noisy pseudo labels, and an uncertainty minimization is applied to
the average probability prediction of the three branches.
$\mathcal{L}_{sup}$ is only for labeled images.} \label{overall}
\end{figure}

\section{Methods}

As illustrated in Fig.~\ref{overall}, the proposed Cross Distillation of Multiple Attentions (CDMA) framework for semi-supervised 
pathological image segmentation consists of three core modules: 1) a tri-branch network MTNet that uses three different attention mechanisms to obtain diverse outputs, 2)
a Cross Decoder Knowledge Distillation (CDKD) module to reduce the effect of noisy pseudo 
labels based on soft supervision, and 3) an average prediction-based uncertainty minimization 
loss to further regularize the predictions on unlabeled images.

\vspace{-2mm}
\subsection{Multi-attention Tri-branch Network (MTNet)}
Attention is an effective network structure design in fully supervised image segmentation~\cite{woo2018cbam,Roy2019}. It can calibrate the feature maps for better performance by paying more attention to the important spatial positions or channels with only a few extra parameters. However, 
it has been rarely investigated in semi-supervised segmentation tasks. 
To more effectively exploit attention mechanisms for semi-supervised pathological image segmentation, our proposed MTNet consists of a shared encoder and three decoder branches that are based on Channel Attention (CA), Spatial Attention (SA) and simultaneous Channel and Spatial Attention (CSA), respectively. The encoder consists of multiple convolutional blocks that are sequentially connected to a down-sampling layer, and each decoder has multiple convolutional blocks that are sequentially connected by an up-sampling layer. For a certain decoder, it uses CA, SA or SCA at the convolutional block at each resolution level to calibrate the features. 

\subsubsection{CA branch}
uses channel attention blocks to calibrate the features in the first decoder. A channel attention block highlights important channels in a feature map and it is formulated as:
\begin{equation}
    F_{c} = F\cdot \sigma\Big(MLP\big(Pool^S_{avg}(F)\big)+MLP\big(Pool^S_{max}(F)\big)\Big)
\end{equation}
Where $F$ represents an input feature map. $Pool^S_{avg}$ and $Pool^S_{max}$ represent average pooling and max-pooling across the spatial dimension, respectively.
$MLP$ and $\sigma$ denote multi-layer perception and the sigmoid activation function respectively. $F_c$ is the output feature map calibrated by channel attention.

\subsubsection{SA branch} leverages spatial attention to highlight the most relevant spatial positions and suppress the irrelevant regions in a feature map. An SA block is:
\begin{equation}
    F_{s} = F\cdot \sigma\Big(Conv\big(Pool^C_{avg}(F) \oplus Pool^C_{max}(F)\big)\Big)
\end{equation}
Where $Conv$ denotes a convolutional layer. 
$Pool^C_{avg}$ and $Pool^C_{max}$ are average and max-pooling across the channel dimension, respectively.  $\oplus$ means concatenation. 

\subsubsection{CSA branch} calibrates the feature maps using a CSA block for each convolutional block. A CSA block consists of a CA block followed by an SA block, taking advantage of  channel and spatial attention simultaneously. 

Due to the different attention mechanisms, the three decoder branches pay attention to different aspects of feature maps and lead to different outputs. 
To further improve the diversity of the outputs and alleviate over-fitting, we add a dropout layer and a feature noise layer $\eta$~\cite{ouali2020semi} before each of the three decoders.
For an input image, the logit predictions obtained by the three branches are denoted as $Z_{CA}$, $Z_{SA}$ and $Z_{CSA}$, respectively. After using a standard Softmax operation, their corresponding probability prediction maps are denoted as $P_{CA}$, $P_{SA}$ and $P_{CSA}$, respectively. 

\vspace{-2mm}
\subsection{Cross Decoder Knowledge Distillation (CDKD)}
\label{sec:CDKD}
Since the three branches have different decision boundaries, using the predictions from one branch as pseudo labels to supervise the others would avoid each branch over-fitting its bias. 
However, as the
predictions for unlabeled training images are noisy and inaccurate, using hard or sharpened pseudo labels~\cite{chen2021semi,wu2022mutual} would strengthen the confidence on incorrect predictions, leading the model to overfit the noise~\cite{muller2019does,xu2020fnkd}. 
To address this problem, we introduce
CDKD to enhance the ability of our MTNet to leverage unlabeled images and 
eliminate the negative impact of noisy pseudo labels.
It forces each decoder 
to be supervised by the other two decoders’ soft predictions.
Following the practice of KD~\cite{hinton2015distilling}, a temperature calibrated Softmax (T-Softmax) is used to soften the probability maps:
\begin{equation}
    \tilde{\mathbf{p}}_c = \frac{exp(\mathbf{z}_c/T)}{\sum_{c}exp(\mathbf{z}_c/T)}
    \label{t-soft}
\end{equation}
where $\mathbf{z}_c$ represents the logit prediction for class $c$ of a pixel, and $\tilde{\mathbf{p}}_c$ is the soft probability value for class $c$. 
Temperature $T$ is a parameter to control the softness of the output probability. Note that $T=1$ corresponds to a standard Softmax function, and a larger $T$ value leads to a softer probability distribution with higher entropy. When $T<1$, Eq.~\ref{t-soft} is a sharpening function.

Let $\Tilde{P}_{CA}$, $\Tilde{P}_{SA}$ and $\Tilde{P}_{CSA}$ represent the soft probability map obtained by T-Softmax for the three branches, respectively. With the other two branches being the teachers, the KD loss for the CSA branch is:
\begin{equation}
    \mathcal{L}_{kd}^{CSA} = \mathbf{KL}(\Tilde{P}_{CSA},\Tilde{P}_{CA}) + \mathbf{KL}(\Tilde{P}_{CSA},\Tilde{P}_{SA})
\end{equation}
where $\mathbf{KL}()$ is the Kullback-Leibler divergence function.
Note that the gradient of $\mathcal{L}_{kd}^{CSA}$ is only back-propagated to the CSA branch, so that the knowledge is distilled from the teachers to the student. Similarly, the  KD losses for the CA and SA branches are denoted as $\mathcal{L}_{kd}^{CA}$ and $\mathcal{L}_{kd}^{SA}$, respectively.
Then, the total distillation loss is defined as:
\begin{equation}
    \mathcal{L}_{cdkd} = \frac{1}{3}(\mathcal{L}_{kd}^{CSA}+\mathcal{L}_{kd}^{CA} + \mathcal{L}_{kd}^{SA})
\end{equation}

\subsection{Average Prediction-based Uncertainty Minimization}
\label{sec:um}
Minimizing the uncertainty (e.g., entropy)~\cite{vu2019advent} has been shown to be an effective regularization for predictions on unlabeled images, which increases the model's confidence on its predictions. However, applying uncertainty minimization to each branch independently  may lead to inconsistent predictions between the decoders where each of them is very confident, e.g., two branches predict the foreground probability of a pixel as 0.0 and 1.0 respectively. To avoid this problem and further encourage  inter-decoder consistency for regularization, we propose an average prediction-based uncertainty minimization:
\begin{equation}
    \mathcal{L}_{um} = -\frac{1}{N}\sum_{i=0}^N\sum_{c=0}^{C} \bar{P}^c_i log(\bar{P}^c_i)
\end{equation}
where $\bar{P}=(P_{CSA}+P_{CA}+P_{SA})/3$ is the average probability map. $C$ and $N$ are the class number and pixel number respectively. $\bar{P}^c_i$ is the average probability for class $c$ at pixel $i$. Note that when $\mathcal{L}_{um}$ for a pixel is close to zero, the average probability for class $c$ of that pixel is close to 0.0 (1.0), which drives all the decoders to predict it as 0.0 (1.0) and encourages inter-decoder consistency.  

Finally, the overall loss function for our CDMA is:
\begin{equation}
    \mathcal{L} = \mathcal{L}_{sup} + \lambda_{1}\mathcal{L}_{cdkd} + \lambda_{2}\mathcal{L}_{um}
    \label{overall_loss}
\end{equation}
where $\mathcal{L}_{sup} = (\mathcal{L}_{sup}^{CSA}+\mathcal{L}_{sup}^{CA}+\mathcal{L}_{sup}^{SA})/3$ is the average supervised learning loss for the three branches on the labeled training images, and the supervised loss for each branch calculates the Dice loss and cross entropy loss between the probability prediction ($P_{CSA}$, $P_{CA}$ and $P_{SA}$) and the ground truth label. 
$\lambda_{1}$ and $\lambda_{2}$ are the weights of 
$\mathcal{L}_{cdkd}$ and $\mathcal{L}_{um}$ respectively.
Note that $\mathcal{L}_{cdkd}$ and $\mathcal{L}_{um}$ are applied on both labeled 
and unlabeled training images.

\begin{center}
\begin{figure}[t]
\includegraphics[width=\textwidth]{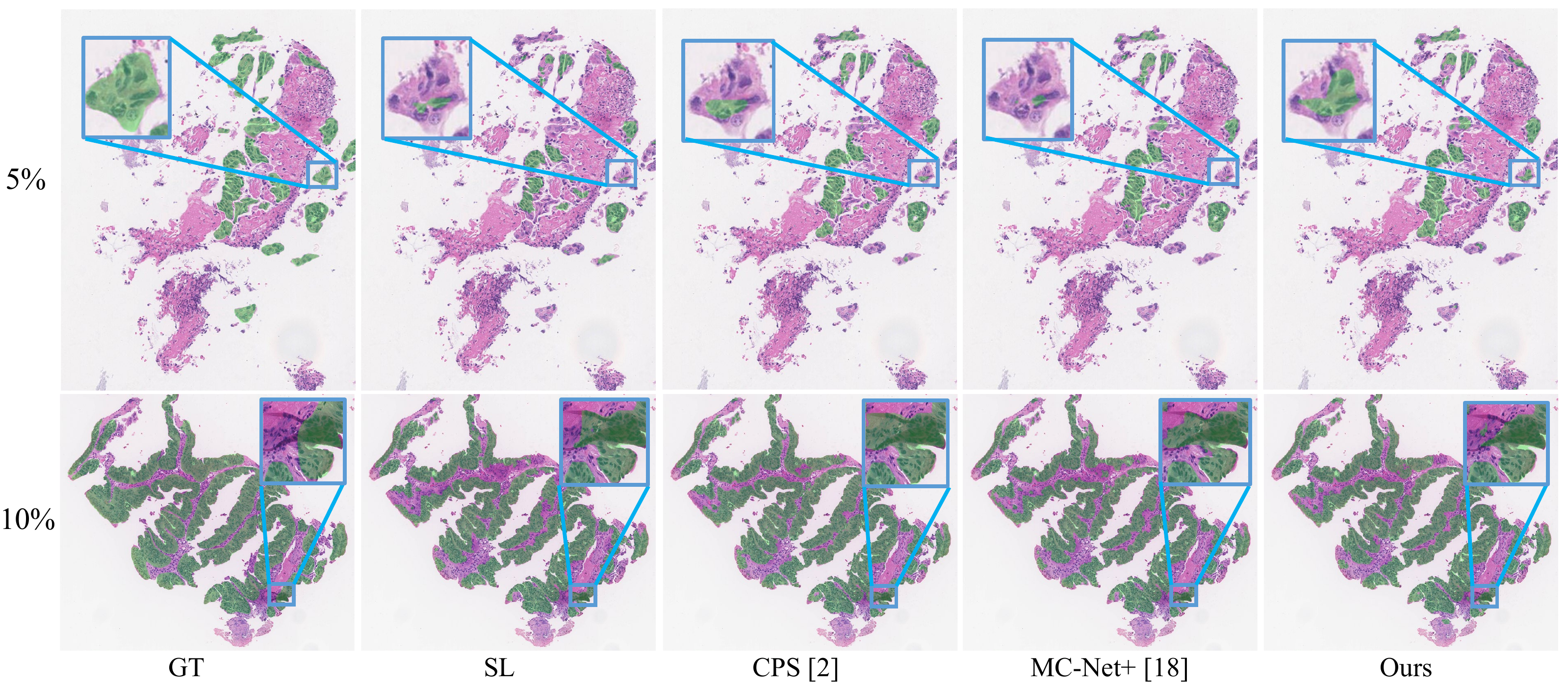}
\caption{Visual comparison between our proposed CDMA with state-of-the-art methods for semi-supervised semantic segmentation of WSIs.
The \textcolor{green}{green} regions are lesions.} 
\label{fig: comparison}
\end{figure}
\end{center}

\vspace{-12mm}
\section{Experiments and Results}
\subsubsection{Dataset and Implementation Details.}
We used the public DigestPath \hyphenation{data-set} 
dataset~\cite{da2022digestpath} for binary segmentation 
of colonoscopy tumor lesions from Whole Slide Images (WSI) in the experiment. The WSIs were 
collected from four medical institutions of $\times$20 magnification (0.475$\mu m$/pixel) 
with an average size of 5000$\times$5000.
We randomly split 130 malignant WSIs into 100, 10, and 20 for
training, validation and testing, respectively.
For SSL, we investigated two annotation ratios: 5\% and 10\%, where only 5 and 
10 WSIs in the training set were taken as annotated respectively.
Labeled WSIs were randomly selected.
For computational feasibility, we cropped the WSIs into patches 
with a size of 256$\times$256.
At inference time for segmenting a WSI, we used a sliding window of size 256$\times$256 with a stride of 192$\times$192.

The CDMA framework was implemented in PyTorch, and all experiments were performed
on one NVIDIA 2080Ti GPU.
MTNet was implemented by extending
DeepLabv3+~\cite{chen2018encoder} into a tri-branch network, where the three decoders were equipped with CA, SA and CSA blocks respectively.  The encoder used a backbone
of ResNet50 pre-trained on ImageNet.
The kernel size of $Conv$ in the SA block is $7\times7$.
SGD optimizer was used for training, with weight decay $5\times10^{-4}$, momentum 0.9 and epoch number 150. The learning rate was initialized to $10^{-3}$ and decayed
by 0.1 every 50 epochs. The hyper-parameter setting was $\lambda_1 = \lambda_2 = 0.1$, $T=10$ 
based on the best results on the validation set.
The batch size was 16 (8 labeled and 8 unlabeled patches).
For data augmentation, we adopted random flipping, random rotation, 
and random Gaussian noise. 
For inference, only the CSA branch was used due to the
similar performance  of the three branches  after converge
and the increased inference time of their ensemble, and no post-processing was used.
Dice Similarity Coefficient (DSC) and Jaccard Index (JI) were used
for quantitative evaluation.

\begin{center}
\begin{table}[t] \centering
\caption{Comparison between  different SSL methods on the DigestPath dataset.
$\ast$ denotes $p$-value < 0.05 (significance level) when comparing the proposed CDMA with the others under t-test hypothesis testing.}
\resizebox{\textwidth}{!}{
\begin{tabular}{c|cc|cc}
\hline
\multirow{2}{*}{Methods} & \multicolumn{2}{c|}{DSC} & 
\multicolumn{2}{c}{Jaccard Index} \\ \cline{2-5} 
& \multicolumn{1}{c|}{5\% labeled} & 10\% labeled & \multicolumn{1}{c|}{5\% labeled}  & 10\% labeled \\ \hline
SL lower bound  & \multicolumn{1}{c|}{64.74$\pm$23.24$^\ast$}  & 68.32$\pm$21.18$^\ast$
& \multicolumn{1}{c|}{52.35$\pm$21.53$^\ast$} & 53.62$\pm$20.32$^\ast$ \\
EM~\cite{vu2019advent} & \multicolumn{1}{c|}{67.09$\pm$24.28$^\ast$}
&70.01$\pm$22.24$^\ast$ & \multicolumn{1}{c|}{54.55$\pm$22.40$^\ast$} 
& 56.96$\pm$21.70$^\ast$ \\
MT~\cite{tarvainen2017mean} & \multicolumn{1}{c|}{67.46$\pm$23.10$^\ast$} & 
70.19$\pm$21.72$^\ast$ & \multicolumn{1}{c|}{54.68$\pm$21.27$^\ast$} 
& 56.38$\pm$21.21$^\ast$ \\
UAMT~\cite{yu2019uncertainty} & \multicolumn{1}{c|}{67.76$\pm$23.44} & 69.64$\pm$22.41$^\ast$ & \multicolumn{1}{c|}{55.16$\pm$22.24} 
& 57.22$\pm$22.25$^\ast$  \\
R-Drop~\cite{wu2021r}  & \multicolumn{1}{c|}{67.22$\pm$24.05$^\ast$}  & 70.37$\pm$23.58$^\ast$ & \multicolumn{1}{c|}{54.70$\pm$22.63$^\ast$} & 
57.39$\pm$22.94$^\ast$ \\
CPS~\cite{chen2021semi} & \multicolumn{1}{c|}{67.71$\pm$22.50$^\ast$} & 70.46$\pm$23.75 & \multicolumn{1}{c|}{54.73$\pm$20.92$^\ast$} 
& 58.67$\pm$23.30\\
HCE~\cite{jin2022semi}& \multicolumn{1}{c|}{67.34$\pm$22.32$^\ast$} & 70.29$\pm$22.62
& \multicolumn{1}{c|}{54.58$\pm$20.37$^\ast$} & 58.04$\pm$21.11 \\
CNN\&Transformer~\cite{luo2022semi}&\multicolumn{1}{c|}{67.66$\pm$25.12} & 70.43$\pm$18.84$^\ast$ & \multicolumn{1}{c|}{55.74$\pm$23.38} & 57.89$\pm$19.48$^\ast$ \\
MC-Net+~\cite{wu2022mutual} & \multicolumn{1}{c|}{67.81$\pm$24.22$^\ast$} & 
70.09$\pm$22.07$^\ast$ & \multicolumn{1}{c|}{55.40$\pm$22.54$^\ast$} & 57.64$\pm$21.80$^\ast$\\
Ours (CSA branch) & \multicolumn{1}{c|}{\textbf{69.72}$\pm$\textbf{22.06}} & \textbf{72.24}$\pm$\textbf{21.21} & \multicolumn{1}{c|}{\textbf{57.09}$\pm$\textbf{21.23}} & 
\textbf{60.17}$\pm$\textbf{21.98} \\ \hline
\multicolumn{1}{c|}{Full Supervision}    & 
\multicolumn{2}{c|}{77.47$\pm$12.49} & 
\multicolumn{2}{c}{64.97$\pm$14.09} \\ \hline
\end{tabular}}
\label{tab: comparison}
\end{table}
\end{center}

\vspace{-12mm}
\subsubsection{Comparison with State-of-the-art Methods.}
Our CDMA was compared with eight existing SSL methods:
1) Entropy Minimization (EM)~\cite{vu2019advent}; 2) Mean Teacher (MT)~\cite{tarvainen2017mean}; 3) Uncertaitny-Aware Mean Teacher (UAMT)~\cite{yu2019uncertainty};
4) R-Drop~\cite{wu2021r} that introduces
a dropout-based consistency regularization between two networks; 
5) CPS~\cite{chen2021semi};
6) Hierarchical Consistency Enforcement (HCE)~\cite{jin2022semi}; 
7) CNN\&Transformer~\cite{luo2022semi} that introduces cross-supervision between CNN and Transformer;
8) MC-Net+~\cite{wu2022mutual} that imposes mutual consistency between multiple slightly different decoders. They were also compared with the lower bound of Supervised Learning (SL)  that only learns from the labeled images.
All these methods used the same backbone of DeepLabv3+~\cite{chen2018encoder} 
for a fair comparison. 

Quantitative evaluation of these methods is shown
in Table~\ref{tab: comparison}.
In the existing methods, MC-Net+~\cite{wu2022mutual}
and CPS~\cite{chen2021semi} showed the best performance for both of 
the two annotation ratios.
Our proposed CDMA achieved a better performance than all the existing methods, with a DSC score of 69.72\% and 72.24\% when the annotation ratio was 5\% and 10\%, respectively. 
Fig.~\ref{fig: comparison} shows a qualitative comparison between different 
methods.
It can be observed that our CDMA yields less mis-segmentation compared with 
CPS~\cite{chen2021semi} and MC-Net+~\cite{wu2022mutual}.

\begin{center}
\begin{table}[t] \centering
\caption{Ablative analysis of our proposed method. 
}
\resizebox{\textwidth}{!}{
\begin{tabular}{l|cc|cc}
\hline
\multirow{2}{*}{Methods} & 
\multicolumn{2}{c|}{Mean DSC}  &  
\multicolumn{2}{c}{Mean JI}   \\ \cline{2-5}
\multicolumn{1}{l|}{} & \multicolumn{1}{c|}{5\% labeled} & 
10\% labeled  & \multicolumn{1}{c|}{5\% labeled} & 10\% labeled \\ \hline
MTNet (Baseline) & \multicolumn{1}{c|}{65.02$\pm$23.94} & 68.61$\pm$22.10 & \multicolumn{1}{c|}{52.59$\pm$22.54} & 55.47$\pm$21.81 \\ 
MTNet + $\mathcal{L}_{cdkd}$ (argmax) & \multicolumn{1}{c|}{68.20$\pm$23.42}& 70.61$\pm$21.03 
& \multicolumn{1}{c|}{55.46$\pm$21.49} & 58.71$\pm$21.23 \\
MTNet + $\mathcal{L}_{cdkd}$ ($T$=1) & \multicolumn{1}{c|}{68.22$\pm$23.55}& 70.32$\pm$21.67 & 
\multicolumn{1}{c|}{55.48$\pm$21.57} & 58.45$\pm$21.32 \\
MTNet + $\mathcal{L}_{cdkd}$ &\multicolumn{1}{c|}{68.84$\pm$22.89}&71.49$\pm$20.74&\multicolumn{1}{c|}{55.92$\pm$21.44} & 59.02$\pm$21.13 \\
MTNet + $\mathcal{L}_{cdkd}$ + $\mathcal{L}_{um}^{\prime}$  
& \multicolumn{1}{c|}{69.11$\pm$23.43}  & 71.56$\pm$22.02 & \multicolumn{1}{c|}{56.57$\pm$21.49} & 59.52$\pm$22.46 \\ 
MTNet + $\mathcal{L}_{cdkd}$ + $\mathcal{L}_{um}$  & 
\multicolumn{1}{c|}{\textbf{69.72}$\pm$\textbf{22.06}} & {72.24}$\pm${21.21} & 
\multicolumn{1}{c|}{\textbf{57.09}$\pm$\textbf{21.23}} & {60.17}$\pm${21.98} \\ \hline
MTNet(dual)~ +$\mathcal{L}_{cdkd}$+$\mathcal{L}_{um}$ &
\multicolumn{1}{c|}{69.49$\pm$22.42}& 71.65$\pm$20.48 
& \multicolumn{1}{c|}{56.96$\pm$21.85} & 59.13$\pm$21.10 \\
MTNet(csa$\times$3)+$\mathcal{L}_{cdkd}$+$\mathcal{L}_{um}$ & \multicolumn{1}{c|}{69.24$\pm$23.57} & 71.50$\pm$20.54
& \multicolumn{1}{c|}{56.93$\pm$22.34} & 59.04$\pm$21.25 \\ 
MTNet(-atten)+$\mathcal{L}_{cdkd}$+$\mathcal{L}_{um}$ 
& \multicolumn{1}{c|}{68.92$\pm$23.42}& 71.37$\pm$20.68 
& \multicolumn{1}{c|}{56.03$\pm$22.13} & 58.81$\pm$21.46 \\
MTNet(ensb)~ +$\mathcal{L}_{cdkd}$+$\mathcal{L}_{um}$ 
& \multicolumn{1}{c|}{69.66$\pm$22.08}& \textbf{72.25}$\pm$\textbf{21.19} 
& \multicolumn{1}{c|}{57.01$\pm$21.25} & \textbf{60.18}$\pm$\textbf{21.98} \\ \hline
\end{tabular}}
\label{tab: ablation}
\end{table}
\end{center}

\vspace{-5mm}
\subsubsection{Ablation Study.}
For ablation study, we set the baseline as using the proposed MTNet with three different decoders for supervised learning from labeled images only. It obtained an average DSC of 65.02\% and 68.61\% under the two annotation ratios respectively. The proposed $\mathcal{L}_{cdkd}$ was compared with two variants: $\mathcal{L}_{cdkd}$ (argmax) and $\mathcal{L}_{cdkd}$ ($T$=1) that represent using hard pseudo labels and standard probability output obtained by Softmax for CDKD respectively. Table~\ref{tab: ablation} shows that our $\mathcal{L}_{cdkd}$ obtained an average DSC of 68.84\% and 71.49\% under the two annotation ratios respectively, and it outperformed $\mathcal{L}_{cdkd}$ (argmax) and $\mathcal{L}_{cdkd}$ ($T$=1), demonstrating that our CDKD based on softened probability prediction is more effective in dealing with noisy pseudo labels. By introducing our average prediction-based uncertainty minimization $\mathcal{L}_{um}$, the DSC was further improved to 69.72\% and 72.24\%  under the two annotation ratios respectively. In addition, replacing our $\mathcal{L}_{um}$ by applying entropy minimization to each branch respectively ($\mathcal{L}'_{um}$) led to a DSC drop by around 0.65\%.

Then, we compared different MTNet variants: 1) MTNet(dual) means a dual-branch structure (removing the CSA branch); 2) MTNet(csa$\times$3) means all the three branches use CSA blocks;  3) MTNet(-atten) means no attention block is used in all the branches; and 4) MTNet(ensb) means using an ensemble of the three branches for inference. Note that all these variants were trained with $\mathcal{L}_{cdkd}$ and $\mathcal{L}_{um}$.
The results in the second section of Table~\ref{tab: ablation} show that using the same structures for different branches, i.e., MTNet(-atten) and MTNet(csa$\times$3), had a lower performance than using different attention blocks, and using three attention branches outperformed just using two attention branches. It can also be found that using CSA branch for inference had a very close performance to MTNet(ensb), and it is more efficient than the later.

\section{Conclusion}
We have presented a novel semi-supervised framework based on Cross Distillation of Multiple Attentions (CDMA)
for pathological image segmentation. 
It employs a Multi-attention Tri-branch network to generate diverse predictions 
based on channel attention, spatial attention, and simultaneous channel and 
spatial attention, respectively.
Different attention-based decoder branches focus on various aspects of feature 
maps, resulting in disparate outputs, 
which is beneficial to semi-supervised learning.
To  eliminate the negative impact of incorrect pseudo labels in training, we 
employ a Cross Decoder Knowledge Distillation (CDKD) to enforce each branch to 
learn from soft labels generated by the other two branches.
Experimental results on a colonoscopy tissue segmentation dataset demonstrated 
that our CDMA outperformed
eight state-of-the-art SSL methods.
In the future, it is of interest to apply our method to multi-class segmentation tasks and pathological images from different organs.

\section{Acknowledgment}
This work was supported by the National Natural Science Foundation of China (62271115).

\newpage
\bibliographystyle{splncs04}
\bibliography{mybib}

\end{document}